\documentclass[sigconf, screen]{acmart}
\renewcommand\footnotetextcopyrightpermission[1]{} 
\AtBeginDocument{%
  \providecommand\BibTeX{{%
    Bib\TeX}}}
\usepackage{enumitem}
\usepackage{graphicx}%
\usepackage{multirow}%

\usepackage{amsmath,amssymb,amsfonts}%
\usepackage{amsthm}%
\usepackage{mathrsfs}%
\usepackage[title]{appendix}%
\usepackage{xcolor}%
\usepackage{textcomp}%
\usepackage{manyfoot}%
\usepackage{booktabs}%
\usepackage{algorithm}%
\usepackage{algorithmicx}%
\usepackage{algpseudocode}%
\usepackage{listings}%
\usepackage{pifont} 
\usepackage{colortbl} 
\usepackage{bm}
\usepackage{cuted}

\AtBeginDocument{%
  \providecommand\BibTeX{{%
    \normalfont B\kern-0.5em{\scshape i\kern-0.25em b}\kern-0.8em\TeX}}}

\setcopyright{none}           
\renewcommand\footnotetextcopyrightpermission[1]{}

\acmConference[Arxiv Preprint]{Under Review}{2026}{}
\acmYear{2026}
\acmDOI{}
\acmISBN{}
\acmPrice{}

\begin{document}

\title{Beyond Waypoints: A Trajectory-Centric Waypointing Paradigm for Vision-Language Navigation}


\author{Haoxiang Shi}
\affiliation{%
  \institution{Harbin Institute of Technology (Shenzhen)}
  \city{Shenzhen}
  \country{China}
}
\affiliation{%
  \institution{Pengcheng Laboratory}
  \city{Shenzhen}
  \country{China}
}
\email{Shihaoxiang1999@gmail.com}

\author{Xiang Deng}
\affiliation{%
  \institution{Harbin Institute of Technology (Shenzhen)}
  \city{Shenzhen}
  \country{China}
}
\email{dengxiang@hit.edu.cn}

\author{Haoyu Zhang}
\affiliation{%
  \institution{Harbin Institute of Technology (Shenzhen)}
  \city{Shenzhen}
  \country{China}
}
\affiliation{%
  \institution{Pengcheng Laboratory}
  \city{Shenzhen}
  \country{China}
}
\email{zhang.hy.2019@gmail.com}

\author{Qiaohui Chu}
\affiliation{%
  \institution{Harbin Institute of Technology (Shenzhen)}
  \city{Shenzhen}
  \country{China}
}
\affiliation{%
  \institution{Pengcheng Laboratory}
  \city{Shenzhen}
  \country{China}
}
\email{qiaohuichu8599@gmail.com}

\author{Yaowei Wang}
\affiliation{%
  \institution{Harbin Institute of Technology (Shenzhen)}
  \city{Shenzhen}
  \country{China}
}
\affiliation{%
  \institution{Pengcheng Laboratory}
  \city{Shenzhen}
  \country{China}
}
\email{wangyw@pcl.ac.cn}

\author{Liqiang Nie}
\affiliation{%
  \institution{Harbin Institute of Technology (Shenzhen)}
  \city{Shenzhen}
  \country{China}
}
\email{nieliqiang@gmail.com}

\renewcommand{\shortauthors}{Haoxiang Shi et al.}


\begin{teaserfigure}
    \includegraphics[width=\textwidth]{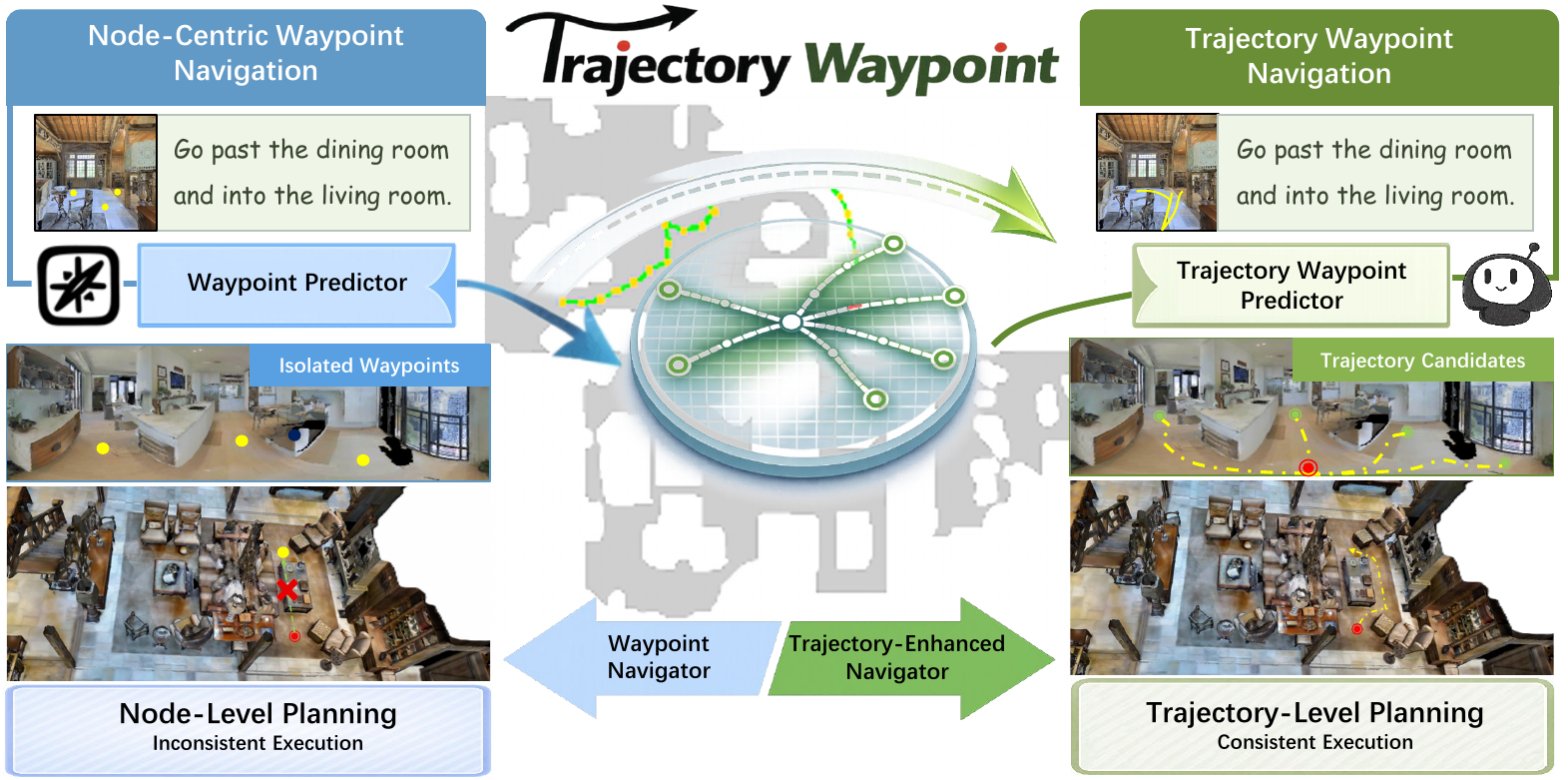}
    \caption{Comparison of navigation paradigms. Traditional node-centric methods (left) predict isolated waypoints, decoupling planning from control and causing inconsistent execution. Conversely, our Trajectory Waypoints paradigm (right) generates continuous trajectory candidates. This enables trajectory-level planning that explicitly evaluates path geometry, tightly coupling high-level planning with low-level execution.}
    \Description{Introduction for Trajectory Waypoint Paradigm.}
    \label{fig1}
\end{teaserfigure}

\begin{abstract}
Vision-Language Navigation in Continuous Environments (VLN-CE) requires agents to follow natural-language instructions while navigating in real-world-like environments.
Most VLN-CE approach\-es adopt a three-stage framework: a waypoint predictor proposes navigable waypoints, and a navigator selects the best waypoint, with a low-level controller executing the movement to it.
However, this decoupled paradigm often leads to unreachable waypoints or inconsistencies between planning and control.
In this work, instead of predicting isolated waypoints, we introduce a novel paradigm called Trajectory Waypoint, which grounds each candidate waypoint in an executable trajectory. 
To realize this, we design a Trajectory Waypoint Predictor formulated as a TSDF-guided diffusion policy, which steers trajectory generation away from obstacles, inherently ensuring the reachability of the predicted waypoints.
We further propose a trajectory-enhanced navigator that injects the associated trajectory as additional information for planning, enabling strict consistency between high-level semantic decisions and low-level execution. 
Extensive experiments on the VLN-CE benchmark show that our Trajectory Waypoint paradigm achieves superior performance over the baselines.
\end{abstract}
\maketitle

\section{Introduction}


Vision-Language Navigation (VLN) is an embodied navigation task in which an agent needs to follow language instructions grounded in observations to reach a specified goal location \cite{anderson2018vision}. The agent continuously perceives its surroundings and makes sequential action decisions under partial observability, requiring it to align linguistic cues with scene understanding and to plan over long horizons in different environments.
Within this setting, waypoints are a fundamental tool for structuring navigation. A waypoint is an intermediate navigational sub-goal that is spatially reachable in the environment and can be used to divide a long trajectory into shorter and more manageable segments. This decomposition provides an explicit interface between high-level planning and low-level control. In VLN, waypoints naturally serve as intermediate targets that guide the agent’s progress through the environment\cite{chen2025affordances}. By converting continuous navigation into a sequence of waypoint-conditioned decisions, an agent can plan locally while maintaining global consistency with the instruction, thereby reducing the search space and mitigating the difficulty of long-horizon decision making in complex scenes.

Vision-Language Navigation in Continuous Environments (VLN-CE) \cite{krantz2020beyond} typically relies on waypoint-based methods operating under a hierarchical framework. First, a waypoint predictor generates a set of node-centric waypoints in the local environment \cite{hong2022bridging}. Second, a high-level navigator selects the optimal waypoint that aligns semantically with the natural language instruction. Finally, a low-level controller drives the agent to execute actions toward the selected target \cite{an2024etpnav}.
Recent waypoint predictors have become increasingly stronger by injecting richer spatial priors into the proposal stage. Typical improvements include leveraging explicit geometry or occupancy cues to better identify traversable regions, and adopting stronger visual representations from large-scale pretraining to improve local navigability estimation \cite{an2024etpnav,shi2025smartway}. Despite these advances, two major issues still remain. First, the candidate distribution remains suboptimal; predicted node-centric waypoints often overlap with obstacles or non-traversable regions, inevitably introducing unreachable targets. Second, the decoupling of waypoint selection and action execution leads to planning–control inconsistencies, where the agent's executed endpoint could deviate from the planned waypoint.

An intuitive alternative to address these issues is to generate executable trajectories directly \cite{wang2025dreamnav}, bypassing the decoupled process of predicting node-centric waypoints and executing them via a separate controller. Recent progress in generative models provides a promising foundation for this approach. Such models have demonstrated a remarkable ability to generate temporally coherent action chunks\cite{DBLP:journals/corr/abs-2410-24164,black2025pi,wei2025ground}. In the navigation domain, works like NoMaD \cite{sridhar2024nomad} have shown that a diffusion-based policy trained on demonstration trajectories can generate collision-free and multi-modal trajectories conditioned on observations, enabling controllable trajectory prediction. Inspired by these advances, we move beyond waypoint-based methods toward a trajectory-centric waypoint paradigm, which naturally alleviates the limitations of hierarchical designs in VLN-CE.

In this work, we propose a novel paradigm called Trajectory Waypoint, which moves beyond conventional node-centric waypoints to generate trajectory candidates. Instead of predicting spatially isolated coordinates, our method generates a diverse set of controllable, collision-free trajectories originating from the agent’s current position. Each candidate thus encapsulates not merely a spatial destination, but an entire executable path, inherently ensuring that every proposed candidate is physically reachable by construction.

To generate these executable trajectory candidates, we propose the Trajectory Waypoint Predictor, formulating it as a diffusion policy \cite{chi2023diffusion} that supports multi-sample generation. To promote safety, we introduce TSDF-based cost guidance \cite{zeng2025navidiffusor} that biases the sampling process away from obstacles, using a local signed-distance field to penalize trajectory samples approaching collision. The resulting guided sampling yields a diverse set of navigable, low-risk trajectories that explore different directions around the agent.

Furthermore, we design a trajectory-enhanced navigator that capitalizes on the rich geometric information embedded in these trajectory candidates. Unlike traditional node-centric waypoints that provide only coarse, point-wise guidance and lack path geometry, our complete trajectories offer dense spatiotemporal context. To fully exploit this advantage, we condition the planning module directly on these trajectories, explicitly encoding both their spatial geometry and temporal structure. This formulation allows the navigator to evaluate not merely the final destination, but specifically how it gets there, enabling tight alignment between the path geometry and fine-grained instruction semantics. 

We comprehensively validate our Trajectory Waypoint paradigm on the VLN-CE benchmark. Crucially, within the VLN-CE scenes, Our Trajectory Waypoint Predictor significantly outperforms existing waypoint predictors in target reachability, achieving a substantially higher \%Open.
Building on these highly reliable trajectory candidates, extensive experiments demonstrate that our overall framework significantly outperforms the baselines on the downstream VLN-CE tasks.

In this work, our main contributions are as follows:
\begin{enumerate}[leftmargin=*]

\item  Unlike existing waypoint-based paradigms for VLN, we propose a trajectory-centric navigation paradigm named Trajectory Waypoint, replacing traditional node-centric waypoint prediction with the generation of executable trajectory candidates. Compared to prior methods, this framework improves sub-goal feasibility and resolves planning-execution inconsistency.
\item  We further develop a trajectory-enhanced navigator that plans among trajectory candidates, rather than isolated node-centric waypoints, by conditioning the navigator on enriched trajectory information. This design enables instruction-conditioned path selection while tightly coupling high-level planning with low-level execution.

\item  We demonstrate that the proposed Trajectory Waypoint paradigm consistently improves performance on the VLN-CE benchmark, yielding robust gains over the baselines.
\end{enumerate}

\section{Related Work}

\subsection{Waypoint-Based Methods in VLN}
Early VLN research predominantly operated within discrete environments, modeling navigation as a node selection problem on predefined connectivity graphs \cite{chang2017matterport3d, hao2020towards, chen2021history,chen2022think, chen2024mapgpt,zhou2024navgpt,zheng2024towards}. As the field transitioned to Vision-Language Navigation in Continuous Environments (VLN-CE), directly transferring these discrete strategies proved unstable due to the extended action horizons required for continuous control \cite{krantz2020beyond, krantz2021waypoint, georgakis2022cross, chen2022weakly}. To bridge this gap, waypoint predictors were introduced \cite{hong2022bridging} to dynamically generate local connectivity graphs, effectively re-discretizing the continuous space. This established the foundation for the waypoint-based methods, allowing high-level planners to select discrete targets that are subsequently executed by low-level controllers.

Recently, Vision-Language Models (VLMs) have been integrated to introduce powerful zero-shot reasoning capabilities to the field \cite{long2024instructnav, qi2025vln, zeng2025janusvln, kim2025test, zhang2026spatialnav, zheng2025efficient,zhou2025same, lin2024correctable}. Despite these semantic advancements, these models remain structurally constrained by the legacy of node-centric waypoints. Lacking fine-grained geometric comprehension, they frequently propose visually salient but physically unreachable sub-goals. Because planning is segregated from execution, this reliance on isolated waypoints inevitably induces a severe planning-control inconsistency, particularly in obstacle-dense environments.
\subsection{Generative Models in VLN}
To bypass the limitations of node-centric waypoints, recent efforts have explored end-to-end generative control paradigms. Driven by advancements in generative models, these frameworks demonstrate that temporal visual context can be directly mapped to low-level control signals, effectively internalizing spatial reasoning within the policy \cite{zhang2024navid, zhang2024uni, wei2025streamvln}. Concurrently, world models and Vision-Language-Action (VLA) models have emerged in the VLN domain \cite{lian2026mapdream, wei2025ground, yao2025navmorph, hu2025astranav, zhang2025embodied,liu2024volumetric}. These approaches utilize VLMs for high-level semantic foresight while employing lightweight generative models to synthesize smooth, collision-free trajectories, thereby resolving the horizon problem and ensuring physical consistency.



Building upon these advancements, our work introduces a novel trajectory-centric waypoint paradigm. Unlike legacy waypoint-based methods that suffer from geometric unreachability, our framework directly predicts fully executable trajectory candidates. This continuous representation fundamentally transforms how the agent perceives spatial transitions. 
By evaluating complete geometric paths rather than isolated coordinates, we tightly align high-level navigation planning with low-level physical control, successfully grounding the interpretability of hierarchical planning in the executable path.

\section{Preliminary}
\subsection{Task Formulation \& Waypoint-Centric Paradigm}

The VLN-CE task is formulated as a Partially Observable Markov Decision Process where an agent navigates a continuous 3D space $\mathcal{C}_{\mathrm{free}}$ guided by a natural language instruction $\mathcal{I}$. At each step $t$, the agent perceives the environment through a panoramic observation $O_t$, consisting of 12 RGB-D image pairs captured at $30^\circ$ intervals following standard protocols:
\begin{equation}
O_t=\{(I_{t, \theta}^{RGB}, I_{t, \theta}^{Depth}) \mid \theta \in \{0^\circ, 30^\circ, \dots, 330^\circ\}\}.
\end{equation}
This observation is then mapped to an atomic action $a_t$ from the discrete action space $\mathcal{A}=\{\texttt{forward}, \allowbreak \texttt{turn\_left}, \allowbreak \texttt{turn\_right}, \allowbreak \texttt{stop}\}$.

To bridge the gap between high-level semantic instructions and low-level continuous execution, existing works predominantly adopt a waypoint-centric paradigm. Inheriting the connectivity-graph structure from legacy discrete environments, this paradigm abstracts navigation into a tractable node selection task. 
Specifically, a trained waypoint predictor discretizes the egocentric panorama into a polar grid and learns a navigability distribution $p_\theta(w_t | s_t)$:
\begin{equation}
p_\theta(w_t^{(k)} | s_t) = \frac{\exp(f_\theta(w_t^{(k)}, s_t))}{\sum_{j} \exp(f_\theta(w_t^{(j)}, s_t))},
\end{equation}
to generate a local set of node-centric waypoints $\mathcal{W}_t$. 
Then a high-level navigator selects the optimal sub-goal $\hat{w}_t \in \mathcal{W}_t$, which is executed by a deterministic low-level controller $\pi_{\mathrm{low}}$ (e.g., a ``Rotate-then-Forward'' heuristic) to translate the selected sub-goal.
\subsection{Limitations of the Waypoint-Centric Paradigm}
Fundamentally bound by the connectivity-graph abstraction, this waypoint-centric paradigm suffers from two critical deficiencies in continuous environments:
\begin{enumerate}[leftmargin=*]
\item \textbf{Geometric Unreachability:} By reducing continuous space to isolated nodes, predictors often ignore complex 3D geometric constraints. A predicted node-centric waypoint is frequently physically invalid (e.g., located inside obstacles or across non-traversable gaps). Formally, defining the physically reachable set as $W_R \subset \mathcal{C}_{\mathrm{free}}$, this point-centric abstraction often yields highly-scored but unreachable targets: 
\begin{equation}
\exists w \in \mathcal{W}_t \text{ s.t. } p_\theta(w | s_t) > \delta \land w \notin W_R .
\end{equation}
Selecting such a sub-goal forces the underlying controller into terminal collisions.
\item \textbf{Planning-Control Inconsistency:} Waypoint-based methods strictly decouple the semantic planning of $\hat{w}_t$ from the physical control $\pi_{\mathrm{low}}$. The planner assumes a direct, unobstructed path to $\hat{w}_t$. However, the reactive low-level controller generates an actual execution trajectory $\mathcal{T}_{\mathrm{exec}}$ that often diverges significantly to avoid obstacles, resulting in 
\begin{equation}
\mathcal{T}_{\mathrm{exec}}(\pi_{\mathrm{low}}, \hat{w}_t) \neq \hat{w}_t.
\end{equation}
This disconnect violates instruction semantics (e.g., swinging wide instead of ``passing near the table''), inevitably leading to catastrophic error accumulation over long horizons.
\end{enumerate}

\section{Trajectory Waypoint}

\begin{figure*}[h]
    \centering
    \includegraphics[width=1.0\textwidth]{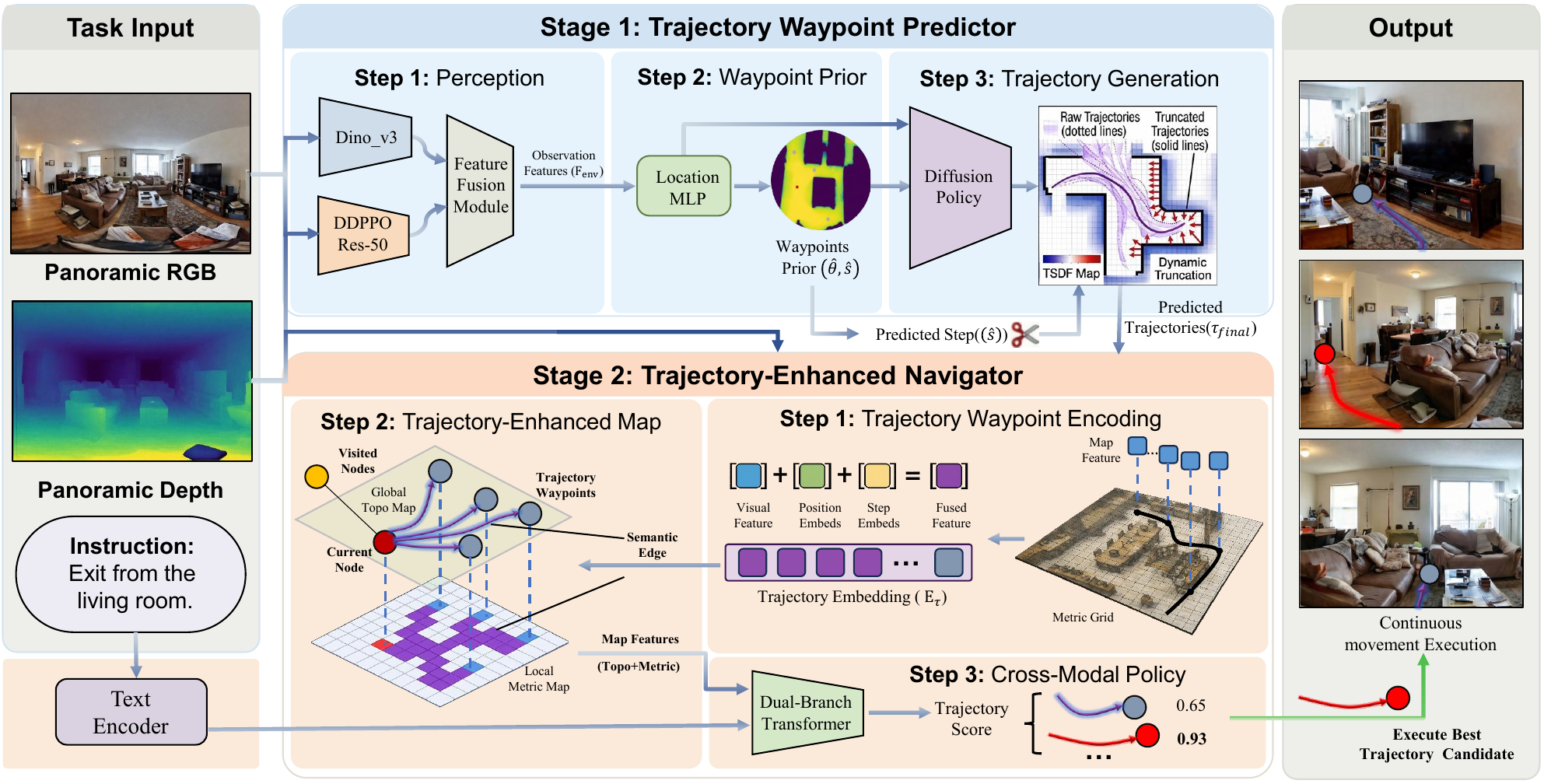}
    \caption{Overview of the proposed Trajectory Waypoint framework. The Trajectory Waypoint Predictor (Stage 1) generates diverse, physically feasible trajectory candidates via environment-guided diffusion. The Trajectory-Enhanced Navigator (Stage 2) then evaluates these continuous paths within a hybrid map, directly coupling high-level semantic reasoning with low-level execution.}\label{fig2}
    \Description{The overview of Trajectory Waypoint.}
\end{figure*}

To address the aforementioned challenges, we propose Trajectory Waypoint, a novel trajectory-centric waypoint paradigm that shifts the atomic unit of VLN from node-centric waypoints to executable trajectory candidates. As illustrated in Figure 2, our framework consists of two stages: a Trajectory Waypoint Predictor (TWP) that generates geometrically feasible trajectory candidates, and a Trajectory-Enhanced Navigator (TEN) that grounds these trajectories in the navigation instruction.

\subsection{Trajectory Waypoint Predictor}
The Trajectory Waypoint Predictor (TWP) is designed to approximate the distribution of feasible trajectory candidates conditioned on the current observation. Formally, we model a trajectory candidate $\tau$ as a sequence of $T$ relative displacements:
\begin{equation}
\tau = \{ \Delta x_{1}, \Delta x_{2}, \dots, \Delta x_{T} \}, \quad \Delta x_{i} \in \mathbb{R}^2,
\end{equation}
where $T$ denotes the prediction horizon. Accordingly, the Trajectory Waypoint is defined as the cumulative sum of these displacements from the current state $x_t$:
\begin{equation}
w_t^{\mathrm{traj}} = x_t + \sum_{i=1}^{T} \Delta x_{i},
\end{equation}
where $x_t$ represents the agent's current position.

\textbf{Observation Encoding:}
We process the panoramic observation $O_t$ using a dual-stream architecture to capture complementary semantic and geometric information. For visual inputs, we employ DINOv3 to extract robust object-level semantic features. In parallel, depth images are encoded by a ResNet-50 pre-trained with DDPPO to extract occupancy and geometric cues. 

To fuse the RGB and depth features, we introduce a Feature Fusion Module inspired by \cite{shi2025smartway}. The fused representations are subsequently processed by a Transformer-based panorama encoder to capture global context, resulting in the final unified observation embedding $F_{\mathrm{env}}^{t} \in \mathbb{R}^{12 \times H}$.

\textbf{Trajectory Waypoint Location Estimation:}
Instead of generating trajectory candidates blindly in the continuous space, we first predict coarse navigational intentions to anchor the subsequent generation process. We employ a lightweight MLP-based predictor that maps the observation embedding $F_{\mathrm{env}}^{t}$ to a joint probability distribution over discretized movement actions:
\begin{equation}
P_{\mathrm{loc}} = \text{Softmax}(\text{MLP}_{\mathrm{loc}}(F_{\mathrm{env}}^{t})) \in \mathbb{R}^{|\Theta| \times |S|},
\end{equation}
where $|\Theta|$ and $|S|$ denote the number of discretized direction bins and step count categories, respectively. From this distribution, we select the top-$K$ candidates to form a seed set:
\begin{equation}
\mathcal{W}_{\mathrm{seed}} = \{ (\hat{\theta}_k, \hat{s}_k) \mid k=1, \dots, K \}.
\end{equation}
Each sampled intention $p_{\mathrm{loc}}^{k}=(\hat{\theta}_k, \hat{s}_k)$ serves as a location prior for generating a corresponding trajectory candidate.

\textbf{Trajectory Waypoint Generation via Environment-Guided Diffusion Policy:}
We formulate the generation of trajectory candidates as a conditional modeling problem, employing a diffusion policy to model the distribution $p_\phi(\tau | S_t^k)$. 
To bridge the discrete location prior with continuous trajectory generation, we construct a condition state $S_t^k$ for each selected candidate $(\hat{\theta}_k, \hat{s}_k)$. Specifically, we retrieve the sector-wise visual feature $F_{\mathrm{target}}^{k}$—aligned with the predicted direction $\hat{\theta}_k$—from the global environment embedding $F_{\mathrm{env}}^{t}$, and fuse it with the positional embedding of the prior:
\begin{equation}
S_t^k = F_{\mathrm{target}}^k \oplus E_{\mathrm{loc}}(\hat{\theta}_k, \hat{s}_k).
\end{equation}
This state $S_t^k$ guides the diffusion process to generate physically feasible paths toward the intended region.

The diffusion process consists of a forward pass that incrementally adds Gaussian noise to a ground-truth trajectory $\tau_0$, and a learnable reverse process that recovers a feasible trajectory from pure noise $\tau_M \sim \mathcal{N}(0, I)$. We employ a denoising network $\epsilon_\phi$ to predict the noise component at each diffusion step $i$. The trajectory is iteratively refined via the following update rule:
\begin{equation}
\tau_{i-1} = \frac{1}{\sqrt{\alpha_i}} \left( \tau_i - \frac{1-\alpha_i}{\sqrt{1-\bar{\alpha}_i}} \epsilon_\phi(\tau_i, i, S_t^k) \right) + \sigma_i \mathbf{z},
\end{equation}
where $i$ denotes the diffusion step from $M$ down to $1$, $\mathbf{z} \sim \mathcal{N}(0, I)$, and $\alpha, \bar{\alpha}, \sigma$ are noise schedule parameters. 
By performing this reverse sampling process once for each of the top-$K$ priors, we generate a set of $K$ diverse trajectory candidates $\{\tau^{(k)}\}_{k=1}^{K}$, each with a fixed prediction horizon $T$, covering the most promising navigational paths.

\textbf{Affordance Constraints for Trajectory Waypoint:}
To rigorously ensure the geometric reachability of the predicted trajectory candidates $\{\tau^{(k)}\}_{k=1}^{K}$, we incorporate a physical validity constraint during the diffusion sampling process. 
Inspired by \cite{zeng2025navidiffusor}, we construct a local Truncated Signed Distance Field (TSDF) map $M_{\mathrm{TSDF}}$ from the current depth observation. We define a differentiable collision cost $\mathcal{J}_{\mathrm{safe}}$ that penalizes trajectory points $p_i$ falling within the truncation distance of an obstacle:
\begin{equation}
\mathcal{J}_{\mathrm{safe}}(\tau) = \sum_{i=1}^{T} \max(0, \epsilon - M_{\mathrm{TSDF}}(p_i)).
\end{equation}
During each denoising step $i$, we apply inference-time guidance by modifying the predicted mean using the gradient of this safety cost:
\begin{equation}
\tau_{i-1}^{\mathrm{guided}} = \tau_{i-1} - \lambda \nabla_\tau \mathcal{J}_{\mathrm{safe}}(\tau).
\end{equation}
This effectively ``pushes'' the generated trajectory away from obstacles via the TSDF gradient, ensuring that the final trajectory $\tau_0$ lies strictly within the reachable set $W_R$.

\textbf{Adaptive Trajectory Truncation:}
The standard diffusion policy generates trajectories with a fixed horizon $T$, which implicitly assumes that endpoints are distributed at rigid intervals. However, diverse navigation scenarios often require flexible action horizons to match local geometric constraints. To mitigate this rigidity, we implement a dynamic pruning mechanism guided by the predicted step count $\hat{s}_k$ from the location prior $p_{\mathrm{loc}}^{k}$. We truncate the generated trajectory to this estimated optimal length:
\begin{equation}
\tau_{\mathrm{final}}^k = \{ \Delta x_1^k, \dots, \Delta x_{\hat{s}_k}^k \}.
\end{equation}
This operation transforms the fixed-length output into variable-length trajectory candidates, ensuring the final physical endpoint ($w_t^{\mathrm{traj}}$) aligns precisely with the environment's affordances.

\subsection{Trajectory-Enhanced Navigator}
Unlike traditional navigators that merely reason about where to go, our Trajectory-Enhanced Navigator (TEN) reasons about how to get there. We achieve this by embedding the generated trajectory candidates into a topo-metric hybrid map, explicitly encoding both path geometry and semantics.

\textbf{Visually-Grounded Trajectory Encoding:}
The generated trajectory candidates $\{\tau^{(k)}_{\mathrm{final}}\}_{k=1}^{K}$ initially lack semantic context. To enable instruction-guided reasoning, we ground these paths in a local metric map $\mathbf{M}_{t} \in \mathbb{R}^{H \times W \times D}$ constructed from projected panoramic visual features. Similar to \cite{an2023bevbert}, panoramic visual features $F_{\mathrm{rgb}}$ are projected into the grid cells of $\mathbf{M}_{t}$ utilizing point clouds derived from the depth map. 

For a candidate trajectory $\tau_{\mathrm{final}} = \{p_0, \dots, p_{\hat{s}}\}$, we retrieve the visual feature $I_i$ at each step coordinate $p_i$. We then fuse this visual context with positional and temporal information via learnable embeddings:
\begin{equation}
h_{i} = I_i + e_{\mathrm{pos}}(p_i) + e_{\mathrm{step}}(i),
\end{equation}
where $e_{\mathrm{pos}}(\cdot)$ and $e_{\mathrm{step}}(\cdot)$ denote the position and time step embedding layers, respectively.
The final trajectory embedding is represented as the sequence $\mathbf{E}_{\tau}= \{h_{0}, h_{1}, \dots, h_{\hat{s}}\}$, where the final hidden state $h_{\hat{s}}$ serves as the trajectory endpoint feature $h_{w}$.

\textbf{Trajectory-Enhanced Map Representation:}
We maintain a topo-metric hybrid map $\mathcal{M}_t = \langle \mathcal{G}_t, \mathcal{V}_{\mathrm{metric}} \rangle$ for comprehensive scene representation. 
Unlike legacy waypoint-based methods that treat map nodes as isolated spatial coordinates, we actively enhance them using the path-dependent features of our trajectory candidates. 

For the topological graph $\mathcal{G}_t$, the candidate endpoints serve as ghost nodes, which represent the potential sub-goals and are connected to the current node via edges defined by $\tau_{\mathrm{final}}$. We use a learnable query vector $q_{\mathrm{traj}}$ to aggregate the sequential trajectory embedding $\mathbf{E}_{\tau}$ into a fixed-dimensional edge representation $e_{\tau}$:
\begin{equation}
e_{\tau} = \text{Attention}(Q=q_{\mathrm{traj}}, K=\mathbf{E}_{\tau}, V=\mathbf{E}_{\tau}).
\end{equation}
This trajectory edge feature explicitly captures the navigational cost and semantics of the path. We then inject it into the visual embedding $v_{\mathrm{node}}$ of the corresponding ghost node: \begin{equation}
\hat{v}_{\mathrm{node}} = v_{\mathrm{node}} + e_{\tau}.
\end{equation}
Simultaneously, we apply a parallel trajectory-to-cell enhancement mechanism to the local metric map $\mathcal{V}_{\mathrm{metric}}$, enriching the dense spatial grid with path-aware semantics.

\textbf{Instruction-Grounded Trajectory Selection:}
To derive the final navigation planning, we feed the trajectory-enhanced hybrid map $\mathcal{M}_t$ and the text-encoded instruction $\mathcal{I}$ into a dual-branch cross-modal Transformer. Specifically, the topological and metric representations are processed using Graph-Aware Self-Attention (GASA) and standard self-attention, respectively, producing the features based on the instruction $N_t = \{n_1, \dots, n_K\}$ and $M_t = \{m_1, \dots, m_K\}$. Subsequently, two Feed-Forward Networks compute a navigability score for each candidate trajectory based on these features. The agent then selects the optimal candidate via a Softmax distribution over these scores.
\begin{equation}
P(w^{\mathrm{traj}}_{t,k} | \mathcal{I}, \mathcal{M}_t) = \text{Softmax}(\text{Score}(n_k, m_k)).
\end{equation}

Crucially, because our framework outputs a fully specified trajectory candidate, the agent directly executes the continuous movement $\tau^{\mathrm{best}}_{\mathrm{final}}$. This tight coupling between high-level semantic planning and low-level control inherently eliminates the planning-execution inconsistency prevalent in legacy waypoint-based methods.

\section{Experiment}

\subsection{Experiment Setup}
In this section, we present a comprehensive evaluation of the proposed Trajectory Waypoint paradigm. We primarily conduct experiments on the VLN-CE benchmark using Habitat simulator.

\subsubsection{Training Details}
\textbf{TWP Training:} A critical priority when training TWP is mitigating geometric overfitting. Previous works typically rely solely on the Matterport3D (MP3D) dataset, which provides only 61 building-scale environments for training. While visually diverse, this limited number of unique room configurations restricts model generalization. To address this, we significantly scale our training data by augmenting MP3D with the Habitat-Matterport 3D (HM3D) dataset, contributing an additional 794 scenes with superior visual fidelity and geometric complexity. Following the data generation protocol in \cite{wang2025navrag}, we sample pre-defined waypoints across HM3D and utilize the Habitat PathFinder to extract continuous demonstration trajectories connecting these points. In total, we constructed a robust dataset comprising over 120,000 waypoint-trajectory pairs to train the TWP.

\textbf{TEN Training:} For the Trajectory-Enhanced Navigator (TEN), we adopt the proven two-stage training paradigm introduced by \cite{an2024etpnav}. The navigator is first pre-trained on the discrete Room-to-Room (R2R)\ dataset\cite{anderson2018vision} to acquire foundational visual-linguistic alignment. Subsequently, we fine-tune the model online within the continuous Habitat simulator using the DAgger (Dataset Aggregation) algorithm, allowing the navigator to effectively adapt to our trajectory-based navigation paradigm.


\subsubsection{Metrics}
\textbf{Predictor Quality Metrics:}
To comprehensively evaluate the generated trajectory candidates, we adopt standard proposal metrics from \cite{shi2025smartway} and incorporate geometric evaluations specifically tailored to continuous paths.
\begin{itemize}[leftmargin=*]
\item $\bm{|\Delta|}$: The absolute difference between the number of predicted candidates and the ground-truth waypoints.
\item \textbf{\%Open}: The percentage of predicted trajectory points that fall strictly within the navigable mesh (free space). A higher \%Open indicates superior geometric feasibility and safety.
\item \textbf{$\bm{d_c}$ (Chamfer Distance):} Measures the average geometric divergence between the predicted trajectory waypoint set and the ground-truth waypoints.
\item \textbf{$\bm{d_h}$ (Hausdorff Distance):} Captures the worst-case geometric error by measuring the maximum distance from any point in the predicted candidate set to the nearest point in the ground truth.
\end{itemize}
\textbf{Navigation Metrics:}
For the downstream VLN-CE task, we report standard evaluation metrics established by \cite{krantz2020beyond}.
\begin{itemize}[leftmargin=*]
\item \textbf{Success Rate (SR)}: The fraction of episodes where the agent successfully halts within 3.0 m of the target location.
\item \textbf{Oracle Success Rate (OSR)}: The fraction of episodes where the agent comes within 3.0 m of the target at any point during its executed trajectory.
\item \textbf{Success weighted by Path Length (SPL)}: Assesses navigation efficiency by penalizing unnecessarily long paths:
\begin{equation}
\text{SPL} = \text{SR} \times \frac{L_{\mathrm{gt}}}{\max(L_{\mathrm{gt}}, L_{\mathrm{pred}})},
\end{equation}
where $L_{\mathrm{gt}}$ and $L_{\mathrm{pred}}$ denote the lengths of the shortest path and the agent's actual path, respectively.
\item \textbf{Collision Rate (CR)}: Defined as the average number of collisions encountered per action step.
\end{itemize}



\subsection{Trajectory Waypoint Predictor}

\begin{table}[th]
\centering
\caption{Performance comparison of waypoint predictors on the scenes from the VLN-CE val-unseen split. Best results are in bold ($\downarrow$: lower is better, $\uparrow$: higher is better).}
\Description{The main experiment for TWP.}
\label{tab:waypoint_predictor}
\begin{tabular}{c l cccc}
\toprule
\multirow{2}{*}{\#} & \multirow{2}{*}{Model} & \multicolumn{4}{c}{VLN-CE Val-Unseen} \\
\cmidrule(lr){3-6}
& & $|\Delta|$ & \%Open $\uparrow$ & $d_c \downarrow$ & $d_h \downarrow$ \\
\midrule
1 & Baseline     & 1.37 & 80.18 & 1.08 & 2.16 \\
2 & U-Net\cite{ronneberger2015u}    & 1.21 & 52.54 & 1.01 & 2.00 \\
3 & RecBERT\cite{hong2022bridging}  & 1.40 & 79.86 & 1.07 & 2.00 \\
4 & ETPNav\cite{an2024etpnav}   & 1.39 & 84.05 & 1.04 & 2.01 \\
5 & SmartWay\cite{shi2025smartway} & 1.41 & 87.26 & 1.03 & 1.96 \\
\midrule
6 & TWP(Ours)     & 1.47 & \textbf{95.84} & \textbf{0.54} & \textbf{1.95} \\
\bottomrule
\end{tabular}
\end{table}

\subsubsection{Main Results in VLN-CE Val-Unseen Scenes.}
Table 1 evaluates the performance of various predictors on the scenes from the VLN-CE val-unseen split. The proposed TWP demonstrates a pronounced advantage over existing waypoint predictors in generating safe and highly actionable targets. Most notably, TWP achieves a state-of-the-art \%Open of 95.84, surpassing the strongest prior baseline by a remarkable absolute margin of 8.58. This substantial gain in reachability confirms that our trajectory candidates can reliably navigate continuous free space without colliding with unmapped obstacles. 
Beyond basic feasibility, TWP exhibits highly precise geometric alignment with optimal human-annotated waypoints. It dramatically reduces the Chamfer Distance ($d_c$) to 0.54, nearly halving the error of prior methods, and attains the lowest Hausdorff Distance ($d_h$) of 1.95.
Collectively, these metrics substantiate that our trajectory-centric waypoint predictor generates candidates that are not only inherently safe but also geometrically precise.

\subsubsection{Ablation Study for TWP}To isolate the driving factors behind our performance, we ablate the core components of the TWP in Table 2.
\textbf{Visual Encoder:} Upgrading the visual backbone from ResNet-50 to DINOv3 yields consistent geometric improvements. Even without TSDF guidance, swapping to DINOv3 increases \%Open from 86.87 to 88.34 and noticeably reduces the spatial alignment error ($d_c$) from 1.05 to 0.87. This confirms that DINOv3's robust, fine-grained feature representations provide a fundamentally superior spatial understanding for baseline trajectory generation.
\textbf{TSDF Guidance (TG):} Explicitly incorporating TG triggers the most dramatic performance leap across all configurations. For our best DINOv3 setup, applying TG elevates \%Open from 88.34 to a peak of 95.84, while drastically cutting $d_c$ to 0.54. This empirically validates the critical role of our inference-time guidance: it actively repels generated trajectory candidates from obstacles and invalid zones, strictly expanding the safety margin and ensuring the physical feasibility of continuous traversal.

\begin{table}[t]
\centering
\caption{Ablation study of the Trajectory Waypoint Predictor.}
\Description{The ablation experiment for TWP.}
\label{tab:waypoint_ablation}
\begin{tabular}{c c | c c c c c}
\toprule
RGB Encoder & TG & \%Open $\uparrow$ & $d_c \downarrow$ & $d_h \downarrow$ \\
\midrule
\multirow{2}{*}{ResNet-50} 
 & \ding{55} & 86.87 & 1.05 & 2.16 \\
 & \ding{51} & 94.96 & 0.64 & 1.96 \\
\midrule
\multirow{2}{*}{Dinov3}    
  & \ding{55} & 88.34 & 0.87 & 2.06 \\\
 & \ding{51} & \textbf{95.84} & \textbf{0.54} & \textbf{1.95} \\
\bottomrule
\end{tabular}
\vspace{-0.5cm}
\end{table}

\begin{figure*}[t]
    \centering
    \includegraphics[width=1.0\textwidth]{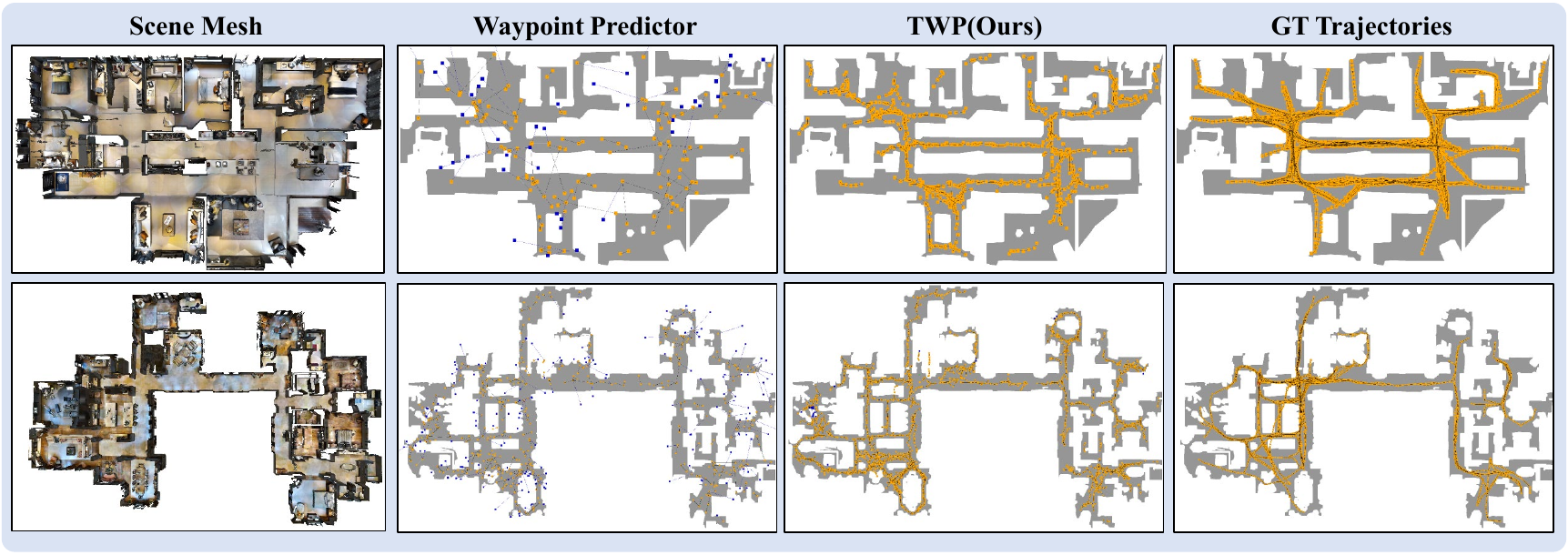}
    \caption{Qualitative results in the scenes from the VLN-CE val-unseen split. Blue points indicate targets generated in inaccessible regions, while yellow points represent physically accessible areas.}
    \Description{Qualitative experiment for TWP in val_unseen scenes.}
\end{figure*}

\subsubsection{Qualitative Experiments for TWP}To rigorously assess the zero-shot generalization ability for our TWP, we densely sample and predict candidates across diverse scenes in the VLN-CE Val-Unseen split. As illustrated in Figure 3, traditional predictors frequently propose invalid node-centric waypoints situated within non-traversable regions (indicated by blue nodes). Furthermore, their reliance on heuristic ``turn-then-forward'' controllers inevitably leads to execution inconsistencies, where the agent's actual action execution diverges from the planned semantic goal. Conversely, guided by the TSDF-based gradient, our TWP successfully generates dense, collision-free trajectory candidates. These generated paths exhibit profound spatial consistency with ground-truth expert demonstrations, inherently guaranteeing kinematic execution fidelity for the downstream navigator. We provide more visualization results for different scenes in the Appendix.

\begin{figure*}[t]
    \centering
    \includegraphics[width=1.0\textwidth]{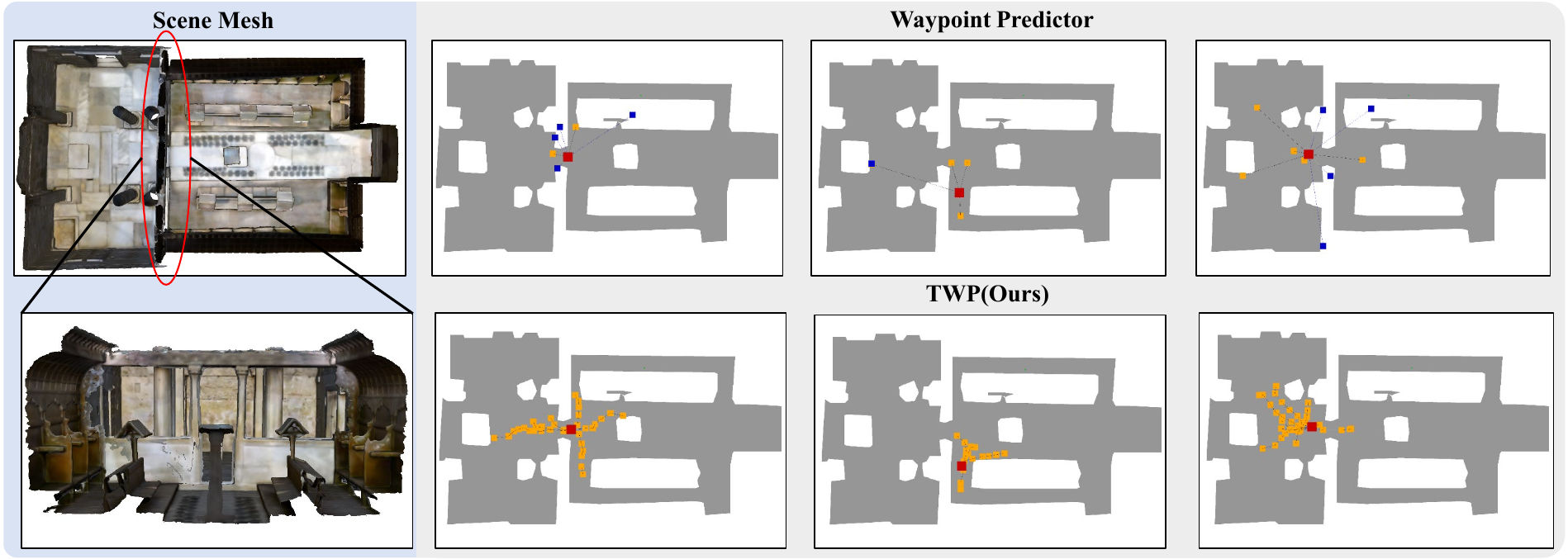}
    \caption{TWP performance under complex geometric constraints. The red point indicates the current location. Blue points indicate targets generated in inaccessible regions, while yellow points represent physically accessible areas.}
    \Description{Complex Environment for TWP.}
\end{figure*}

To further demonstrate the reliability of TWP under complex geometric constraints, Figure 4 isolates a challenging ``visible but unreachable'' scene, a compact church environment bisected by a low wall. Because the region behind the wall falls within the agent's panoramic visual stream but is physically inaccessible, traditional predictors lacking intrinsic 3D comprehension consistently propose unfeasible sub-goals that penetrate the solid boundary, inevitably causing terminal collisions. In stark contrast, our TWP intrinsically integrates spatial TSDF priors during the stochastic diffusion phase. This continuous gradient guidance actively exerts a repulsive spatial force, steering the trajectory candidates strictly away from the impenetrable barrier. This confirms that our approach can reliably yield high-quality, physically executable paths even when subjected to highly deceptive spatial constraints.

\subsection{VLN-CE Downstream Task}\label{subsec9}
\begin{table}[t]
\centering
\caption{Comparison with state-of-the-art methods on the R2R-CE Val-Seen and Val-Unseen splits. * indicates VLM-based methods. $^{\dagger}$ denotes baselines utilizing the waypoint predictor from \cite{hong2022bridging}.}
\label{tab:merged_full_results}

\setlength{\tabcolsep}{1.3pt}

\resizebox{\columnwidth}{!}{%
\begin{tabular}{l cccc cccc}
\toprule
\multirow{2}{*}{Method} & \multicolumn{4}{c}{R2R-CE Val-Seen} & \multicolumn{4}{c}{R2R-CE Val-Unseen} \\
\cmidrule(lr){2-5} \cmidrule(lr){6-9}
& NE $\downarrow$ & OSR $\uparrow$ & SR $\uparrow$ & SPL $\uparrow$ & NE $\downarrow$ & OSR $\uparrow$ & SR $\uparrow$ & SPL $\uparrow$ \\
\midrule
Navid*\cite{zhang2024navid} & 5.62 & 51.5 & 43.5 & 38.4 & 5.72 & 49.2 & 41.9 & 36.5 \\
Uni-Navid*\cite{zhang2024uni} & 4.40 & 64.5 & 59.1 & 55.4 & 5.28 & 53.3 & 47.0 & 42.7 \\
MonoDream*\cite{wang2026monodream} & - & - & - & - & 5.45 & 61.5 & 55.8 & 49.1 \\
MapNav*\cite{zhang2025mapnav} & - & - & - & - & 4.93 & 53.0 & 39.7 & 37.2 \\
Dynam3D*\cite{wang2025dynam3d} & - & - & - & - & 5.34 & 62.1 & 52.9 & 45.7 \\
NaVILA*\cite{cheng2024navila} & 5.05 & 63.8 & 56.4 & 51.5 & 5.22 & 62.5 & 54.0 & 49.0 \\
\midrule
Sim2Sim$^{\dagger}$\cite{krantz2022sim} & 4.67 & 61.0 & 52.0 & 44.0 & 6.07 & 52.0 & 43.0 & 36.0 \\
HPN+DN\cite{krantz2021waypoint} & 5.48 & 53.0 & 46.0 & 43.0 & 6.31 & 40.0 & 36.0 & 34.0 \\
Ego$^2$-Map\cite{hong2023learning} & - & -    & -    & -  & 4.94 & 59.0 & 52.0 & 46.0 \\
CM$^2$\cite{georgakis2022cross} & 4.81 & 58.3  & 52.8 & 41.8  & 6.23 & 41.3 & 37.0 & 30.6 \\
VLN$\circlearrowright$BERT$^{\dagger}$\cite{hong2021vln} & 5.74 & 59.0 & 50.0 & 44.0 & 5.74 & 53.0 & 44.0 & 39.0 \\
GridMM$^{\dagger}$\cite{wang2023gridmm}  & 4.21 & 69.0 & 59.0 & 51.0 & 5.11 & 61.0 & 49.0 & 41.0 \\
DWalker$^{\dagger}$\cite{wang2023dreamwalker} & 5.53 & 66.0 & 59.0 & 48.0 & 5.54 & 59.0 & 49.0 & 44.0 \\
ETPNav$^{\dagger}$\cite{an2024etpnav} & 3.95 & 72.0 & 66.0 & 59.0 & 4.71 & 65.0 & 57.0 & 49.0 \\
BEVBert$^{\dagger}$ \cite{an2023bevbert} & \underline{3.77} & \underline{73.0} & \underline{68.0} & \underline{60.0} & \textbf{4.57} & \underline{67.0} & \underline{59.0} & 50.0 \\
Energy$^{\dagger}$\cite{liu2024vision} & 3.90 & \underline{73.0} & \underline{68.0} & 59.0 & \underline{4.69} & 65.0 & 58.0 & 50.0 \\
\midrule
\rowcolor{gray!20} Ours & \textbf{3.75} & \textbf{74.6} & \textbf{68.8} & \textbf{60.2} & \textbf{4.57} & \textbf{68.1} & \textbf{60.3} & \textbf{51.4} \\
\bottomrule
\end{tabular}
}
\vspace{-0.5cm}
\end{table}

\subsubsection{Main Results in R2R-CE}Table 3 presents a comprehensive performance comparison on the R2R-CE benchmark. We evaluate across both Val-Seen and Val-Unseen splits to rigorously assess domain-specific learning and zero-shot generalization. Notably, the baselines are categorized into traditional small-scale models and recent large-parameter models.

As shown in Table 3, our method establishes a new state-of-the-art on the challenging R2R-CE Val-Unseen split. By replacing traditional node-centric waypoints with geometrically grounded trajectory candidates, our agent attains the highest OSR of 68.1, SR of 60.3 and SPL of 51.4, which highlights that our approach not only navigates to the target reliably but does so through optimal, succinct paths without redundant exploration.
Furthermore, results on the Val-Seen split reinforce the robust learning capacity of our model. Benefiting from the tight coupling of high-level planning and low-level execution, our method achieves the absolute highest SR of 68.8, alongside a peak SPL of 60.2 and an Oracle Success Rate (OSR) of 74.6. Notably, our approach strictly dominates both traditional baselines relying on legacy waypoint predictors and recent VLM-based architectures. This empirically validates the inherent superiority of our trajectory-centric waypoint paradigm in continuous domains.

\begin{table}[t]
\centering
\caption{Ablation study of our Trajectory Waypoint framework on the R2R-CE Val-Unseen split.}
\label{tab:ablation_study}
\begin{tabular}{l ccccc}
\toprule
\multirow{2}{*}{Method} & \multicolumn{5}{c}{R2R-CE Val-Unseen} \\
\cmidrule(lr){2-6}
& NE $\downarrow$ & OSR $\uparrow$ & SR $\uparrow$ & SPL $\uparrow$ & CR $\downarrow$\\
\midrule
Ours                      & \textbf{4.54} & \textbf{68.1} & \textbf{60.3} & \textbf{51.4} & \textbf{0.004}\\
\quad w/o TEN   & 4.68 & 66.8 & 59.3 & 50.3 & 0.004\\
\quad w/o TSDF            & 4.67 & 67.1 & 59.2 & 50.7 & 0.05 \\
\quad w/o TWP   & 4.75 & 65.2 & 58.7 & 49.1 & 0.12\\
\bottomrule
\end{tabular}
\vspace{-0.5cm}
\end{table}
\subsubsection{Ablation Study for Waypoint Trajectory Framework}To systematically validate our structural design, Table 4 details an ablation study on the R2R-CE Val-Unseen split. 
Replacing our predictor with a standard baseline (w/o TWP) severely degrades the SR from 60.3 to 58.7 and SPL from 51.4 to 49.1, accompanied by a massive surge in the CR. This confirms that our geometrically feasible trajectory candidates are the primary driver for safe, collision-free navigation and overall performance.
Furthermore, substituting our Trajectory-Enhanced Navigator with a traditional map fusion module (w/o TEN) noticeably drops SR to 59.3 and SPL to 50.3, while leaving the CR largely unaffected since TEN operates purely at the high-level semantic planning stage. This gap highlights a critical flaw in legacy systems: fundamentally bound by the connectivity-graph's point-centric waypoint navigation paradigm, they merely extract isolated coordinates and fail entirely to exploit the dense continuous context embedded within our trajectories.
Finally, removing the inference-time TSDF guidance (w/o TSDF) causes a consistent decline across all metrics. Without this continuous gradient guidance acting as the physical constraint, a portion of the generated paths inevitably drifts into unreachable regions, directly increasing the CR and impairing overall reachability.
\begin{figure}[t]
    \centering
    \includegraphics[width=0.47\textwidth]{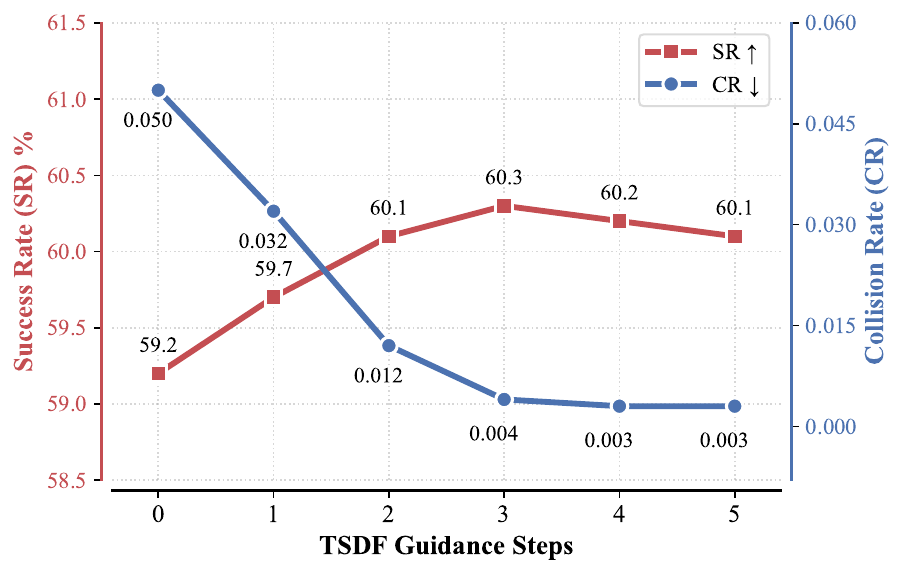}
    \caption{Impact of TSDF-guide steps on navigation performance.}
    \Description{Complex Environment for TWP.}
\vspace{-0.5cm}
\end{figure}

\subsubsection{Effect of TSDF-Guide Steps in the Denoising Process}
During the trajectory generation phase of our TWP, we incorporate a TSDF-guide into the diffusion denoising process to steer the candidate paths away from obstacles. To fully investigate the impact of the number of guidance steps, we conduct the ablation study by applying the TSDF-guide strictly for the first $n$ steps out of the total $k$ denoising steps.
As shown in Figure 5, increasing the guidance steps consistently lowers the CR, proving its effectiveness in steering trajectories away from obstacles. However, the SR peaks at 3 steps and subsequently declines. This indicates a critical trade-off: while moderate guidance improves trajectory safety, excessive guidance over-constrains the generative process, disrupting the diffusion model's original denoising distribution and ultimately causing sub-optimal navigation performance.
\section{Conclusion}
In this paper, we introduce Trajectory Waypoint, a trajectory-centric paradigm for Vision-Language Navigation that resolves the unreachability and planning-control disconnect inherent in legacy waypoint-based methods. By replacing isolated waypoints with continuous, collision-free trajectories generated by our Trajectory Waypoint Predictor, and planning over these candidates via the Trajectory-Enhanced Navigator, we strictly align high-level planning with low-level execution. Extensive experiments on the R2R-CE benchmark confirm our approach significantly outperforms the baselines, providing a physically scalable foundation for robust embodied navigation.
\bibliographystyle{ACM-Reference-Format}
\bibliography{sample-base}

\clearpage
\nobalance
\appendix
\pagenumbering{arabic} 

\begin{strip}
\centering
{\Huge\textbf{Beyond Waypoints: A Trajectory-Centric Waypointing Paradigm for Vision-Language Navigation}}
\vspace{0.5em} \\
\centering
{\huge{Supplementary Material}}
\vspace{0.5em} \\
\vspace{0.5em}
\end{strip}







In the supplementary material, we provide additional details on Trajectory Waypoint Predictor (TWP) training data collection, hyperparameter settings, and more qualitative results of the Trajectory Waypoint framework.

\section{Trajectory Data Collection}
\label{app:traj_collection}

To train the Trajectory Waypoint Predictor (TWP), we construct a safety-aware trajectory supervision dataset within the MatterPort3D (MP3D) and Habitat-Matterport 3D (HM3D) environment. 
Driven by the connectivity graph, navigation inherently follows a point-centric paradigm characterized by straight-line connections between graph nodes. However, these linear paths often pass dangerously close to obstacles, yielding suboptimal and potentially unsafe supervision. 
To mitigate this, we refine the raw rollouts using a Truncated Signed Distance Field (TSDF)-based adjustment, pushing trajectories toward higher-clearance regions.

\paragraph{Point-Centric Trajectory Generation.}
For each scan, we construct a connectivity graph, from which we derive directed source-target node pairs.
For each source node, we synthesize panoramic depth observations to build a local 2D TSDF cost map centered at the source pose. Subsequently, for each valid source-target pair, we simulate a raw trajectory by repeatedly executing forward actions toward the target direction. If a collision is detected during this rollout, the sample is immediately discarded. Collision-free paths are retained for subsequent clearance refinement.

\paragraph{TSDF-Based Clearance Refinement.}
Let $\mathcal{T}=\{\mathbf{p}_1,\dots,\mathbf{p}_N\}$ denote a raw, collision-free straight-line trajectory segment. To increase obstacle clearance, we fix the endpoints ($\mathbf{p}_1$ and $\mathbf{p}_N$) and iteratively update the intermediate waypoints. Given the TSDF cost map $C$, each intermediate point is shifted along the normalized ascent direction of $C$:
\begin{equation}
\mathbf{p}_i \leftarrow \mathbf{p}_i + \eta \cdot \frac{\nabla C(\mathbf{p}_i)}{\|\nabla C(\mathbf{p}_i)\|_2}, \quad i=2,\dots,N-1.
\end{equation}
To prevent excessive path distortion, we apply a length constraint to the adjusted trajectory. If the length of the updated trajectory exceeds $1+\lambda$ times the original raw trajectory length $L_0$, the current adjustment step is rejected, and the step size $\eta$ is halved. The detailed procedure is summarized in Algorithm 1.

\begin{algorithm}[H]
\caption{TSDF-Adjusted Trajectory Collection}
\label{alg:tsdf_traj_collection}
\begin{algorithmic}[1]
\Require Connectivity graph $\mathcal{G}=(\mathcal{V},\mathcal{E})$, simulator $\mathcal{M}$, TSDF parameters $(K,\eta_0,\lambda)$
\Ensure Safety-aware trajectory set $\mathcal{D}$

\State $\mathcal{D} \gets \emptyset$

\ForAll{scan $s$ in dataset}
    \State Initialize simulator $\mathcal{M}$ with scan $s$
    \ForAll{source node $v_a \in \mathcal{V}$}
        \State \textcolor{blue}{\textbf{$(C,\text{meta}) \gets \textsc{BuildTSDF}(\mathcal{M},v_a)$}}
               \Comment{\textcolor{gray}{\small Construct local cost map}}
        \ForAll{neighbor $v_b$ of $v_a$}
            \State $\mathcal{T}_{\mathrm{raw}} \gets \textsc{ForwardRollout}(\mathcal{M},v_a,v_b)$
            \If{collision in $\mathcal{T}_{\mathrm{raw}}$}
                \State \textbf{continue}
            \EndIf
            
            \State $\mathcal{T} \gets \mathcal{T}_{\mathrm{raw}}$
            \If{$|\mathcal{T}| \ge 3$ \textbf{and} $C \neq \varnothing$}
                \State $L_0 \gets \textsc{Length}(\mathcal{T}_{\mathrm{raw}})$
                \State $\eta \gets \eta_0$
                
                \ForAll{iteration $k \in \{1,\dots,K\}$}
                    \State $\mathcal{T}' \gets \mathcal{T}$
                    \ForAll{index $i \in \{2,\dots,|\mathcal{T}|-1\}$}
                        \State $(x_i,z_i) \gets \textsc{WorldToMap}(\mathbf{p}_i,\text{meta})$
                        \If{out of map bounds}
                            \State $\mathcal{T}' \gets \mathcal{T}_{\mathrm{raw}}$
                            \State \textbf{break}
                        \EndIf
                        \State \textcolor{blue}{\textbf{$\mathbf{g} \gets \nabla C(x_i,z_i)$}}
                               \Comment{\textcolor{gray}{\small Compute TSDF gradient}}
                        \If{$\|\mathbf{g}\|_2 > 10^{-6}$}
                            \State \textcolor{blue}{\textbf{$\mathbf{p}_i' \gets \mathbf{p}_i' + \eta \cdot \frac{\mathbf{g}}{\|\mathbf{g}\|_2}$}}
                                   \Comment{\textcolor{gray}{\small Push toward higher clearance}}
                        \EndIf
                    \EndFor
                    
                    \If{$\textsc{Length}(\mathcal{T}') > (1+\lambda)L_0$}
                        \State \textcolor{blue}{\textbf{$\eta \gets 0.5\eta$}}
                               \Comment{\textcolor{gray}{\small Reject over-stretched update}}
                    \Else
                        \State \textcolor{blue}{\textbf{$\mathcal{T} \gets \mathcal{T}'$}}
                               \Comment{\textcolor{gray}{\small Accept update}}
                        \State $\eta \gets 0.7\eta$
                               \Comment{\textcolor{gray}{\small Decay step size}}
                    \EndIf
                \EndFor
            \EndIf
            \State $\mathcal{D} \gets \mathcal{D} \cup \{(v_a, v_b, \mathcal{T})\}$
        \EndFor
    \EndFor
\EndFor

\State \Return $\mathcal{D}$
\end{algorithmic}
\end{algorithm}

\paragraph{Horizon-Aware Trajectory Stitching.}
While the TSDF-adjusted segments ensure local safety, individual point-to-point transitions are often shorter than the action prediction horizon required by our diffusion policy. To address this, we perform trajectory stitching to construct extended, horizon-aligned supervision. Starting from an initial adjusted segment $(v_a \rightarrow v_b)$, we iteratively append valid subsequent segments. To maintain kinematic smoothness and prevent abrupt turns, the selection of the next segment is governed by an angle constraint. Let $\mathbf{u}$ denote the direction of the current segment and $\mathbf{v}$ represent a candidate subsequent direction. We uniformly sample from candidates whose turning angle satisfies $\theta(\mathbf{u},\mathbf{v}) \le \theta_{\max}$ (empirically set to $60^\circ$). If no candidate meets this criterion, we select the one with the minimum turning angle. Ultimately, these seamlessly stitched, long-horizon trajectories serve as expert candidates, providing high-quality, safety-aware supervision for training the TWP.

\begin{figure*}[!t]
    \centering
    \includegraphics[width=0.82\textwidth]{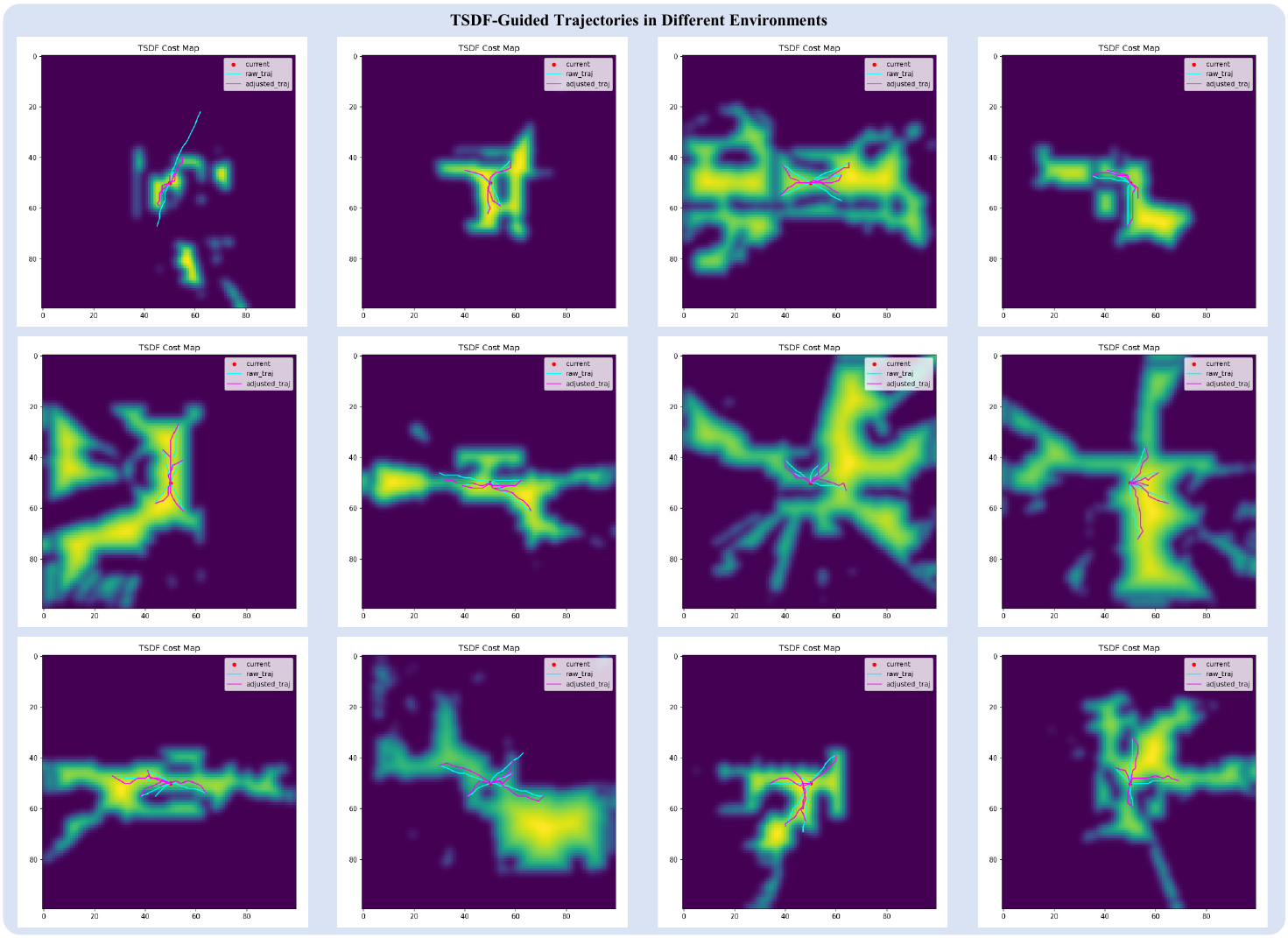}
    \caption{Visualization of TSDF-guided trajectory refinement.}
    \Description{Qualitative experiment for tsdf guide in TWP.}
\end{figure*}

\section{Hyperparameter Settings}

\paragraph{TWP Key Hyperparameters.}
We detail the core hyperparameters for our TWP and the TSDF-guided denoising process. Specifically, each trajectory candidate predicted by the TWP spans a horizon of 12 future actions. At each navigation step, the TWP samples up to 5 directional candidates based on the current panoramic observation. For the TWP prior, the angular and radial discretization strides are configured to 120 and 12, respectively. During the diffusion process, the model undergoes 10 denoising iterations. Regarding the TSDF guidance mechanism, following [43], we set the weight of the TSDF cost term for gradient computation to 0.003, with an update step size of 1.0 applied to the denoised outputs. This guidance is injected at every step for a maximum of 3 denoising iterations. Intuitively, the prediction horizon defines the temporal extent of the trajectory candidates, the prior parameters dictate the spatial discretization granularity, and the TSDF configurations govern the strength and frequency of the obstacle-aware geometric guidance.

\section{Qualitative Analysis of TSDF Guidance}

Figure 1 illustrates the qualitative impact of our TSDF guidance on trajectory generation. In narrow and cluttered environments, unguided predictions ($\texttt{raw\_traj}$) frequently intersect with obstacles due to an inherent straight-line bias. The introduction of TSDF guidance ($\texttt{adjusted\_traj}$) successfully steers the trajectories away from high-cost regions, ensuring they remain strictly within safe, navigable free space. Although this geometric compliance incurs a marginal increase in overall path length, it successfully eliminates collision risks. This confirms that the TSDF guidance effectively prioritizes kinematic safety over aggressive but risky planning.

\section{Additional Visualizations of Trajectory Candidates Generation}

To further demonstrate the robust trajectory generation capabilities of our TWP, Figure 2 presents extended qualitative comparisons across the remaining scenes in the VLN-CE val-unseen split. These supplementary visualizations consistently corroborate our primary findings: regardless of spatial complexity, the TWP successfully avoids generating invalid nodes in non-traversable regions. Instead, it reliably produces continuous, collision-free trajectory candidates that perfectly align with expert demonstrations, thereby ensuring strict execution fidelity across the entire val-unseen set.

\begin{figure*}[t]
    \centering
    \includegraphics[width=0.85\textwidth]{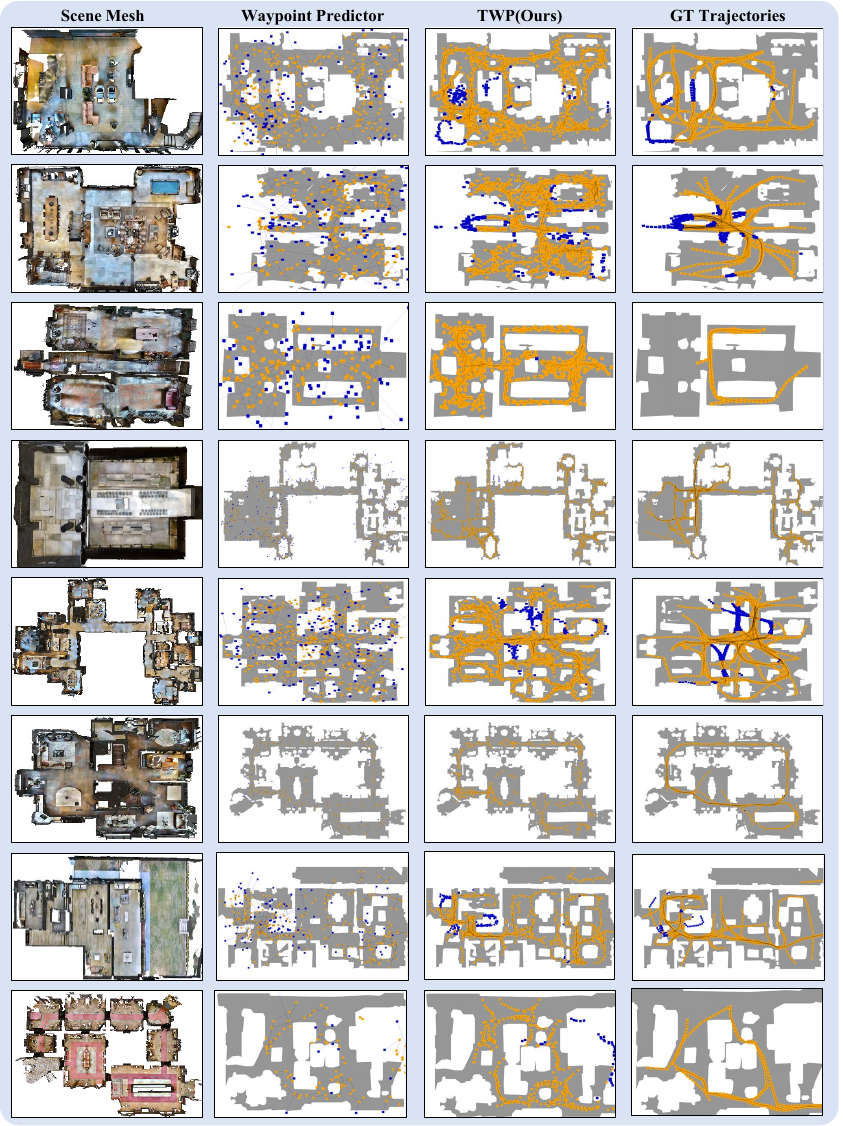}
    \caption{Supplementary visualizations for the VLN-CE val-unseen split. Blue points indicate targets generated in inaccessible regions, while yellow points represent physically accessible areas.}
    \Description{Qualitative experiment for TWP in val\_unseen scenes.}
\end{figure*}

\end{document}